\documentclass{article}

\usepackage[final, nonatbib]{nips_2016}

\usepackage[utf8]{inputenc} 
\usepackage[T1]{fontenc}    
\usepackage{hyperref}       
\usepackage{url}            
\usepackage{booktabs}       
\usepackage{amsfonts}       
\usepackage{nicefrac}       
\usepackage{microtype}      
\usepackage{float}
\usepackage{graphicx}
\usepackage{subcaption}
\usepackage{color}
\usepackage{longtable}
\usepackage{multirow}
\usepackage{supertabular}

\title{CyLKs: Unsupervised Cycle Lucas-Kanade Network for Landmark Tracking}

\author{
  Xinshuo Weng \\
  Carnegie Mellon University\\
  \texttt{xinshuow@andrew.cmu.edu} \\
  \And
  Wentao Han \\
  Carnegie Mellon University\\
  \texttt{whan1@andrew.cmu.edu} \\
}

\begin{document}

\maketitle

\begin{abstract}

Across a majority of modern learning-based tracking systems, expensive annotations are needed to achieve state-of-the-art performance. In contrast, the Lucas-Kanade (LK) algorithm works well without any annotation. However, LK has a strong assumption of photometric (brightness) consistency on image intensity and is easy to drift because of large motion, occlusion, and aperture problem. To relax the assumption and alleviate the drift problem, we propose CyLKs, a data-driven way of training Lucas-Kanade in an unsupervised manner. CyLKs learns a feature transformation through CNNs, transforming the input images to a feature space which is especially favorable to LK tracking. During training, we perform differentiable Lucas-Kanade forward and backward on the convolutional feature maps, and then minimize the re-projection error. During testing, we perform the LK tracking on the learned features. We apply our model to the task of landmark tracking and perform experiments on datasets of THUMOS and 300VW. 
\end{abstract}


\section{Introduction}
Landmark tracking, also known as key-point tracking, is one of the hot areas in computer vision for decades. The quality of the landmark tracking lays the foundation for improving the performance of many other vision tasks, such as face recognition, blendshape modeling, face animation, face reenactment and human action recognition, etc. For example, in face animation and reenactment, 2D landmarks can be used as the control points (fixed constraints) while deforming one mesh to another. In action recognition, one can define the human skeleton as a set of landmarks and classify the action based on the motion of human skeleton.

With the advent of convolutional neural networks (CNNs), there are many modern data-driven tracking systems trained in a supervised manner achieving state-of-the-art performance \cite{Fan2017, Danelljan2016, Valmadre2017, Zajc2016}. The main problem with these methods is that extensive training annotations are needed, which is very expensive. By contrast, the optical flow methods like Lucas-Kanade algorithm does not need any annotation and works well in general cases. But Lucas-Kanade algorithm has the limitation on images with a large variation of illumination changes, aperture problem, occlusion, etc. 

To overcome this, we propose the CyLKs, which is a trainable Lucas-Kanade network. After training on a large amount of video data, the CyLKs is expected to alleviate the problems of illumination changes, aperture problem, etc. Different from the existing work in \cite{Lin2016, Wang2017, ChangChun2017}, which also proposes to combine Lucas-Kanade algorithm with convolution neural networks. CyLKs can be trained in an unsupervised manner without using any human annotation, and thus can leverage a very large amount of unlabeled video data and have better generalization capability.

\begin{figure}[t]
    \centering
    \includegraphics[height=0.4\linewidth]{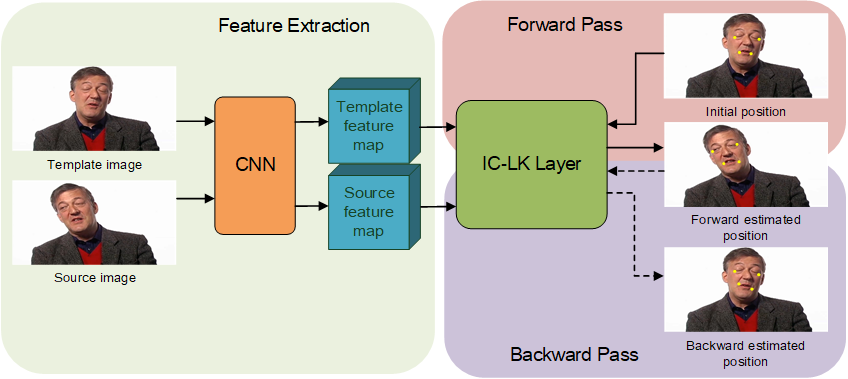}
    \caption{The overview of proposed Cycle Lucas-Kanade Network (CyLKs). }
    \label{fig:cycle-ic-lk-network}
\end{figure}

Specially, CyLKs learns a feature representation that is favorable to LK tracking. This is achieved by: i) Extract features from source and template image using a CNN; ii) Given initial point positions $\textbf{x}_T$ in the template image, tracking forward (forward pass) to estimate positions $\textbf{x}_I$ in the input image; iii) Tracks backward (backward pass) given the estimated points in forward pass $\textbf{x}_I$ to the points $\textbf{x}_T'$ in the template image; iv) Calculate Euclidean loss between initial points $\textbf{x}_T$ and backward tracked points $\textbf{x}_T'$, then pass the gradients back to update CNN parameters. The overall architecture is shown in Figure \ref{fig:cycle-ic-lk-network}.

\noindent\textbf{Contributions.} 1) CyLKs can be trained without using any human annotation, and thus can learn variations from a large number of unlabeled videos. 2) CyLKs can learn a feature transformation, which can improve the performance of Lucas-Kanade based trackers.

\vspace{-0.2cm}
\section{Related Work}
\vspace{-0.2cm}


\noindent\textbf{Unsupervised Tracking.} Before the explosion of deep learning techniques, most unsupervised trackers can be classified into three categories: direct methods, feature-based methods and a hybrid of direct and feature-based methods.

Direct methods, mostly based on Lucas-Kanade algorithm \cite{baker2004lucas}, operate on pixel intensity to estimate the motion between images. Such methods are computationally efficient and have been proved to achieve competitive results in tasks of SLAM \cite{engel2014lsd, alismail2016photometric} and visual odometry \cite{forster2014svo}. However, direct methods assume photometric consistency across frames and are thus not robust to variations of illumination changes, occlusion, and out-of-plane motion.

Instead of working on raw images, feature-based methods extract robust features and estimate motion by matching feature descriptors between images. Robust features such as SIFT \cite{lowe2004distinctive} and ORB \cite{rublee2011orb} are usually used. Without assuming photometric consistency, feature-based methods are more robust to illumination changes. However, the performance of feature-based methods heavily rely on the localization capabilities and matching the accuracy of the features. 

Recently, a hybrid of direct and feature-based methods are explored in \cite{alismail2016robust, antonakos2015feature, bristow2016defense}, which apply direct approaches on the robust features. Unfortunately, although these approaches are proved to have a significant improvement from the direct methods, especially with the improved quality of data (e.g., higher resolution and frame rate), they are still limited to the presence of large motion. 

\noindent\textbf{Supervised Tracking.} With the superior representation capabilities of the CNNs, many CNN-based tracking methods outperform the unsupervised tracking methods. Held et al. \cite{held2016learning} proposed GOTURN which applies a deep regression network to predict object locations based on deep features.
\cite{Nam2015} proposed a classification-based multi-domain tracker, which try to separate the domain-independent information
from domain-specific one, to capture shared representations to some extent.
C-COT \cite{CCOT} introduced the concept of multi-resolution fusion and continuous domain learning for the visual tracking system to achieve accurate sub-pixel feature point tracking.
ECO \cite{Danelljan2016} proposed a factorized convolution operator to reduce the number of parameters and an efficient model update strategy, and achieve significant improvement in both speed and robustness.
\cite{Wang2016} designed a two-stream CNN to handle drastic appearance change and distinguish target object from its similar distracters during tracking.
\cite{Feichtenhofer2017} set up a CNN architecture for simultaneous detection and tracking, and introduced the correlation features to represent object co-occurrences across time to aid tracking.

However, CNN-based trackers usually work in a specific domain and have limited generalization capabilities. 
Also, the performance of such trackers drops significantly on the sequences which have a different appearance from the training data. 
To achieve good generalization, a large number of high-quality annotations are necessary, which turns out to be very expensive.

\vspace{-0.2cm}
\section{Approach} \label{gen_inst}

\subsection{Inverse Compositional Lucas-Kanede (IC-LK) Layer} \label{sec:ic-lk-layer}
Motivated by \cite{ChangChun2017}, the IC-LK layer performs the inverse compositional Lucas-Kanade algorithm \cite{baker2004lucas}. 

\noindent\textbf{LK Algorithm.} Given a template image $I_T$ and an input image $I_I$, we first extract the image features with a CNN $\phi (\cdot)$ to obtain feature maps $F_T$ and $F_I$. Starting from an initial point $\textbf{x}_T$ in the template image, the IC-LK algorithm tries to estimate translation parameters $\textbf{p}$ by minimizing
$$E(\textbf{p}) = \frac{1}{2} ||F_{T}(\textbf{x}_T)-F_{I}(\textbf{x}_{T}+\textbf{p})||^2,$$ 
where $F_T(\textbf{x}_T)$ is a feature patch of $F_T$ centered at $\textbf{x}_T$. This nonlinear objective is then optimized in an iterative manner. Starting from an initial translation $\textbf{p}$, taking the first order Taylor expansion of $F_I(\textbf{x}_T + \textbf{p} + \Delta \textbf{p})$, the translation update term $\Delta \textbf{p}$ is estimated by minimizing 
$$E(\Delta \textbf{p}) = \frac{1}{2} || F_I(\textbf{x}_T + \textbf{p}) - F_T(\textbf{x}_T) + \nabla F_I(\textbf{x}_T + \textbf{p}) \Delta \textbf{p}||^2,$$ where $\nabla F_I(\textbf{x}_T)$ is the gradient patch of $F_T$ centered at $\textbf{x}_T$. The derivative of $E$ w.r.t $\Delta \textbf{p}$ is then given by 
$$\frac{\partial E}{\partial \Delta \textbf{p}} = [\nabla F_I(\textbf{x}_T + \textbf{p})]^T[F_I(\textbf{x}_T + \textbf{p}) + \nabla F_I(\textbf{x}_T + \textbf{p})\Delta \textbf{p} - F_T(\textbf{x}_T)],$$ 
setting this term to zeros gives the least square solution of $\Delta \textbf{p}$ as 
$$\Delta \textbf{p} = H^{-1} [\nabla F_I(\textbf{x}_T + \textbf{p})]^T [F_T(\textbf{x}_T) - F_I(\textbf{x}_T + \textbf{p})],$$ 
where $H = [\nabla F_I(\textbf{x}_T + \textbf{p})]^T \nabla F_I(\textbf{x}_T + \textbf{p})$ is the Hessian matrix of $F_I$ at $(\textbf{x}_T + \textbf{p})$. $\textbf{p}$ is then updated by $\textbf{p} \leftarrow \textbf{p} + \Delta  \textbf{p}$.\\

\noindent\textbf{IC-LK Algorithm.}
Unlike the feature map gradients which can be pre-computed, $H$ have to be computed at each iteration. To avoid such computation, the inverse-compositional method is applied to LK algorithm by transforming the objective function to
$$E(\Delta \textbf{p}) = \frac{1}{2} || F_T(\textbf{x}_T + \Delta \textbf{p}) - F_I(\textbf{x}_T + \textbf{p})||^2 \approx \frac{1}{2} || F_T(\textbf{x}_T) + \nabla F_T(\textbf{x}_T) \Delta \textbf{p} - F_I(\textbf{x}_T + \textbf{p})||^2,$$
which gives the least square solution of $\Delta \textbf{p}$ as
$$\Delta \textbf{p} = H^{-1} [\nabla F_T(\textbf{x}_T)]^T [F_I(\textbf{x}_T + \textbf{p}) - F_T(\textbf{x}_T)],$$
where $H = [\nabla F_T(\textbf{x}_T)]^T \nabla F_T(\textbf{x}_T)$ is the Hessian matrix of $F_T$ at $\textbf{x}_T$. Note that now $H$ can be pre-computed and is fixed throughout the optimization process. Then $\textbf{p}$ is updated by $\textbf{p} \leftarrow \textbf{p} - \Delta  \textbf{p}$ at each iteration. After convergence, the forward predicted point locations in the input image is obtained as $\textbf{x}_I = \textbf{x}_T + \textbf{p}$.

\begin{figure}[ht] 
    \centering
    \includegraphics[width=1.0\linewidth]{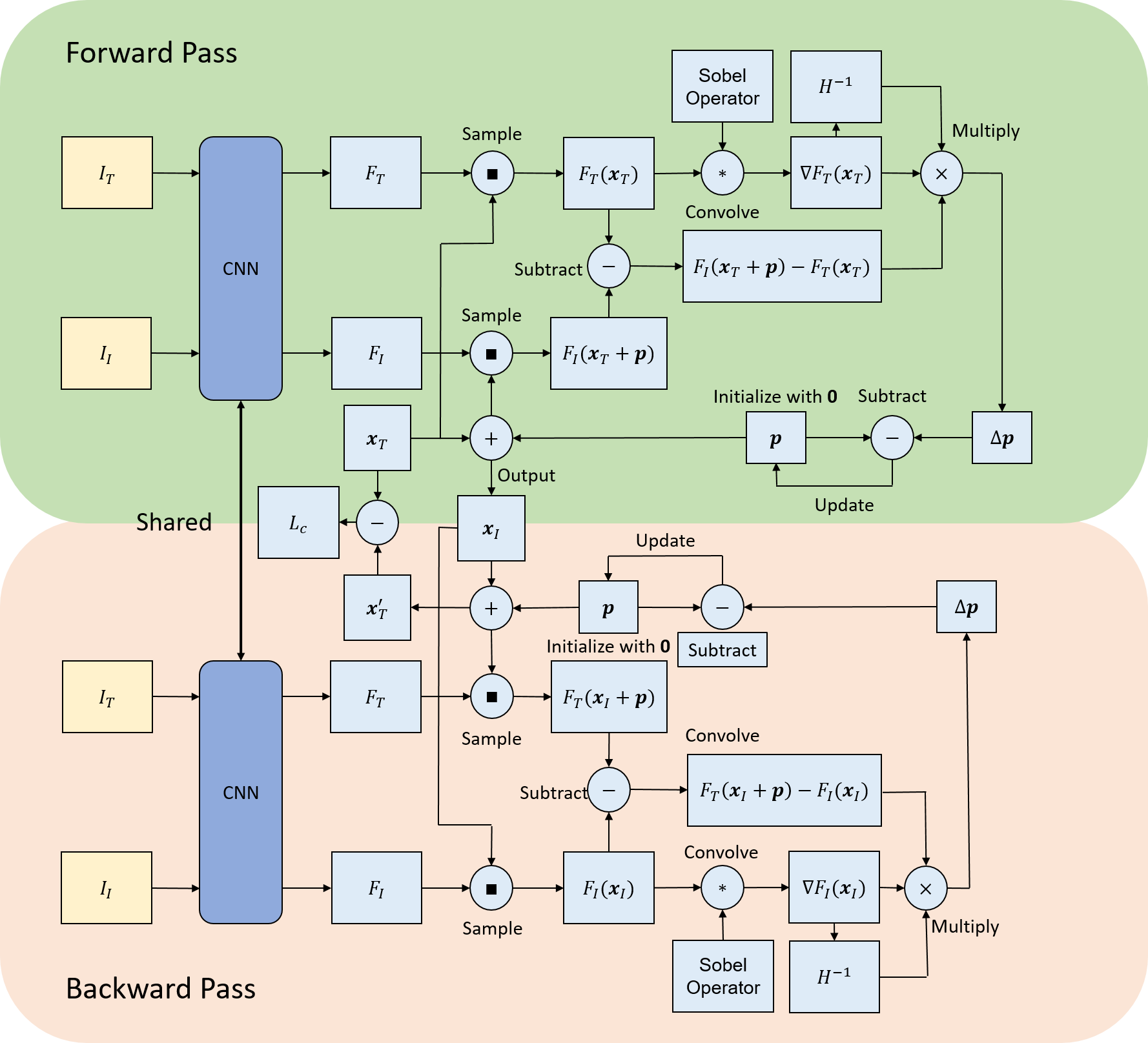}
    \caption{Computation Graph of Cycle IC-LK Network}
    \label{fig:ic-lk-layer}
\end{figure} \label{fig:computation-graph}

\subsection{Cycle IC-LK Network}\vspace{-0.2cm}
Similarly, we apply the algorithm described in \ref{sec:ic-lk-layer} for a backward pass to track $\textbf{x}_I$ back to the template image and obtain $\textbf{x}_T'$. Then the parameters of $\phi (\cdot)$ are updated by minimizing the patch loss $L_p$ and the cycle poss $L_c$, which is illustrated in \ref{sec:loss-function}. Since $\textbf{x}_T$ and $\textbf{x}_T'$ are parameterized by $F_T$ and $F_I$, this approach is end-to-end differentiable and we can propogate the error back to update the learnable parameters. Details of the derivation are described in \ref{sec:back-prop}.

\subsection{Loss Function}\vspace{-0.2cm} \label{sec:loss-function}

\noindent\textbf{Cycle Loss.} The cycle loss $L_c = ||\textbf{x}_T-\textbf{x}_T'||^2$ is the implicit criterion of the tracker's performance. As previously mentioned, our approach tries to learn the trainable parameters in an unsupervised manner by minimizing the cycle loss.

\noindent\textbf{Patch Loss.} Similar to the objective function of IC-LK algorithm, a desirable representation of an image should approximate the photometric consistency assumption, i.e. the two feature patches centered at $x_{T}$ and $x_{I}$ should be similar. We enforce this by introducing a patch loss $L_p = ||F_{T}(\textbf{x}_T)-F_{I}(\textbf{x}_I)||^2$, where $F_{T}(\textbf{x}_T)$ is a patch of feature map $F_{T}$ centered at $\textbf{x}_{T}$.

\noindent\textbf{Overall Loss.} The overall loss is simply the combination of the cycle loss and the patch loss: $L=L_c + \lambda L_p$, where $\lambda$ is factor balancing the two loss terms.

\subsection{Full Derivation} \label{sec:back-prop}

The computation graph of Cycle IC-LK Network is shown in Figure \ref{fig:computation-graph}. In this section, the full derivation of Cycle IC-LK network is given, following the computation flow as shown in the computation graph. 

Starting from an initial point $\textbf{x}_{T}$ in $F_{T}$, suppose $\textbf{p}$ is updated $n$ times to get the final point $\textbf{x}_{I}$, during which a series of temporary points $\{\textbf{x}_1, \textbf{x}_2, \cdots , \textbf{x}_{n-1}\}$ and corresponding updates in warp parameters $\{\Delta \textbf{p}_1, \Delta\textbf{p}_2, \cdots, \Delta \textbf{p}_{n}\}$ are obtained, then we have $\textbf{p} = -\sum_{i=1}^{n} \Delta \textbf{p}_{i}$, where
$$\Delta \textbf{p}_{i} = H^{-1}[\nabla F_T(\textbf{x}_T)]^T[F_I(\textbf{x}_T+\textbf{p}_{i-1})-F_T(\textbf{x}_T)],$$
in which $H = [\nabla F_T(\textbf{x}_T)]^T \nabla F_T(\textbf{x}_T)$ is the Hessian matrix of $F_T$ at $\textbf{x}_T$ and $\textbf{p}_{i-1}$ is the warp parameters after the $(i-1)$th update. Then we further have
$$\textbf{x}_I = \textbf{x}_T + \textbf{p} = \textbf{x}_T - H^{-1}[\nabla F_T(\textbf{x}_T)]^T\sum_{i=1}^{n}[F_I(\textbf{x}_T+\textbf{p}_{i-1}) - F_T(\textbf{x}_T)]$$
Similarly, for the backward pass, suppose the $\textbf{p}$ is updated $m$ times to convergence and $\{\Delta \textbf{p}_1', \Delta \textbf{p}_2', \cdots, \Delta \textbf{p}_m'\}$ are obtained, we can derive
$$\textbf{x}_T' = \textbf{x}_I - (H')^{-1}[\nabla F_I(\textbf{x}_I)]^T\sum_{j=1}^{m}[F_T(\textbf{x}_I+\textbf{p}_{j-1}') - F_I(\textbf{x}_I)],$$
where $H' = [\nabla F_I(\textbf{x}_I)]^T \nabla F_I(\textbf{x}_I)$. Combining the two equations, we can derive the cycle loss as
$$L_c = || H^{-1}[\nabla F_T(\textbf{x}_T)]^T\sum_{i=1}^{n}[F_I(\textbf{x}_T+\textbf{p}_{i-1}) - F_T(\textbf{x}_T)] + (H')^{-1}[\nabla F_I(\textbf{x}_I)]^T\sum_{j=1}^{m}[F_T(\textbf{x}_I+\textbf{p}_{j-1}') - F_I(\textbf{x}_I)] ||^2$$
To propagate the error back to the CNN, we need to compute the gradient of the feature maps $F_T$ and $F_I$, which can be obtained by summing up the gradients of all the feature patches that parameterize $\textbf{x}_T'$ at the corresponding locations of $F_T$ and $F_I$. This requires computing the derivative of $L_c$ w.r.t all feature patches, including template patches $F_T(\textbf{x}_T)$ and $F_I(\textbf{x}_I)$ and non-template patches $F_I(\textbf{x}_T + \textbf{p}_i)$ and $F_T(\textbf{x}_I + \textbf{p}_j')$, which is a complicated computation flow. In practice, the back-propagation is done by the automatic differentiation functionality of PyTorch \cite{Pytorch}.

In addition, in \ref{sec:loss-function} we introduced patch loss to enforce photometric consistency in the feature space, the derivative of which can be simply computed as 
$$\frac{\partial L_p}{\partial F_I(\textbf{x}_I)} = F_I(\textbf{x}_I) - F_T(\textbf{x}_T), \hspace{2ex} \frac{\partial L_p}{\partial F_T(\textbf{x}_T)} = F_T(\textbf{x}_T) - F_I(\textbf{x}_I)$$
Finally, the gradients of $F_T$ and $F_I$ is obtained by summing up the gradients of all feature patches at the locations from which the patches are sampled. The feature map gradients are then back-propagated to the CNN to update the learnable parameters.

\section{Dataset} \label{headings}

\noindent\textbf{THUMOS 2015 Dataset.}
A large video dataset for action recognition, consisting of 13,320 trimmed videos for training, 2,104 untrimmed videos for validation and 5,613 untrimmed videos for testing. The resolution varies from $320\times 240$ to $1280\times 720$. No landmark annotation is provided.

\begin{figure}[ht]
    \centering
    \includegraphics[width=0.8\linewidth]{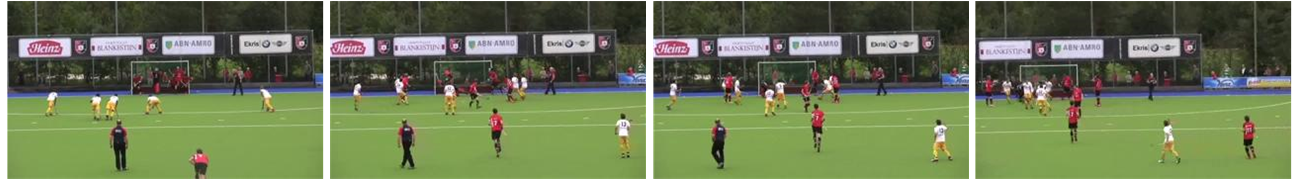}
    \caption{Example sequence from THUMOS 2015 dataset}
    \label{fig:thumos_example}
\end{figure}
    
\noindent\textbf{300-VW.}
A video dataset for face alignment, consisting of 114 videos with each approximately one minute in length. 68 facial landmarks with semantic interpretation are annotated across all the videos.

\begin{figure}[ht]
    \centering
    \includegraphics[width=0.8\linewidth]{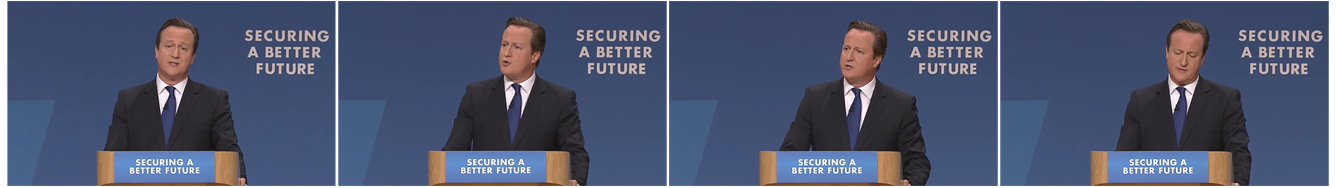}
    \caption{Example sequence from 300-VW dataset}
    \label{fig:300vw_example}
\end{figure}


\begin{figure}[t]
    \centering
    \includegraphics[width=0.2\linewidth]{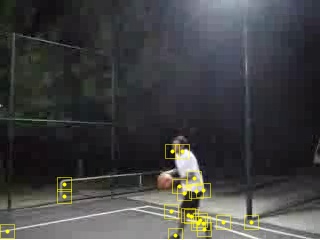}
    \includegraphics[width=0.2\linewidth]{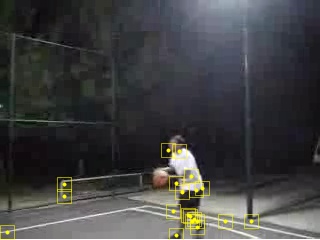}
    \includegraphics[width=0.2\linewidth]{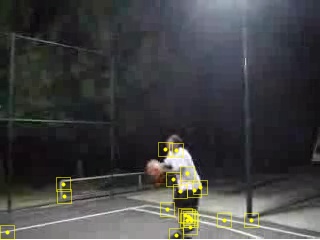}
    \includegraphics[width=0.2\linewidth]{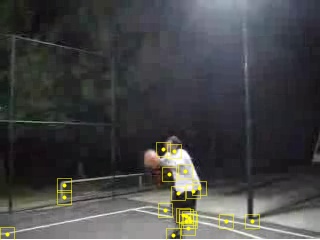}
    \caption{Qualitative results on a sample sequence from the THUMOS dataset.}
    \label{fig:results_thumos}
\end{figure}

\begin{figure}[t]
    \centering
    \includegraphics[width=0.22\linewidth, trim={15cm 13cm 15cm 0cm}, clip]{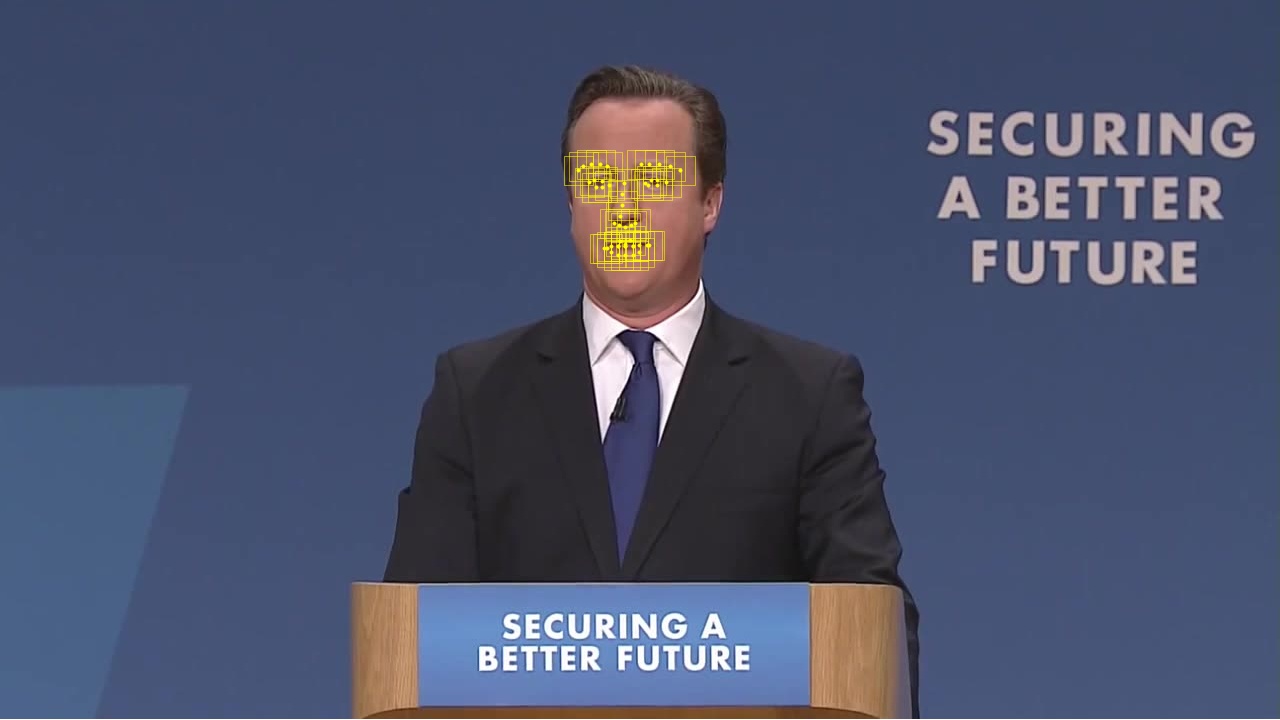}
    \includegraphics[width=0.22\linewidth, trim={15cm 13cm 15cm 0cm}, clip]{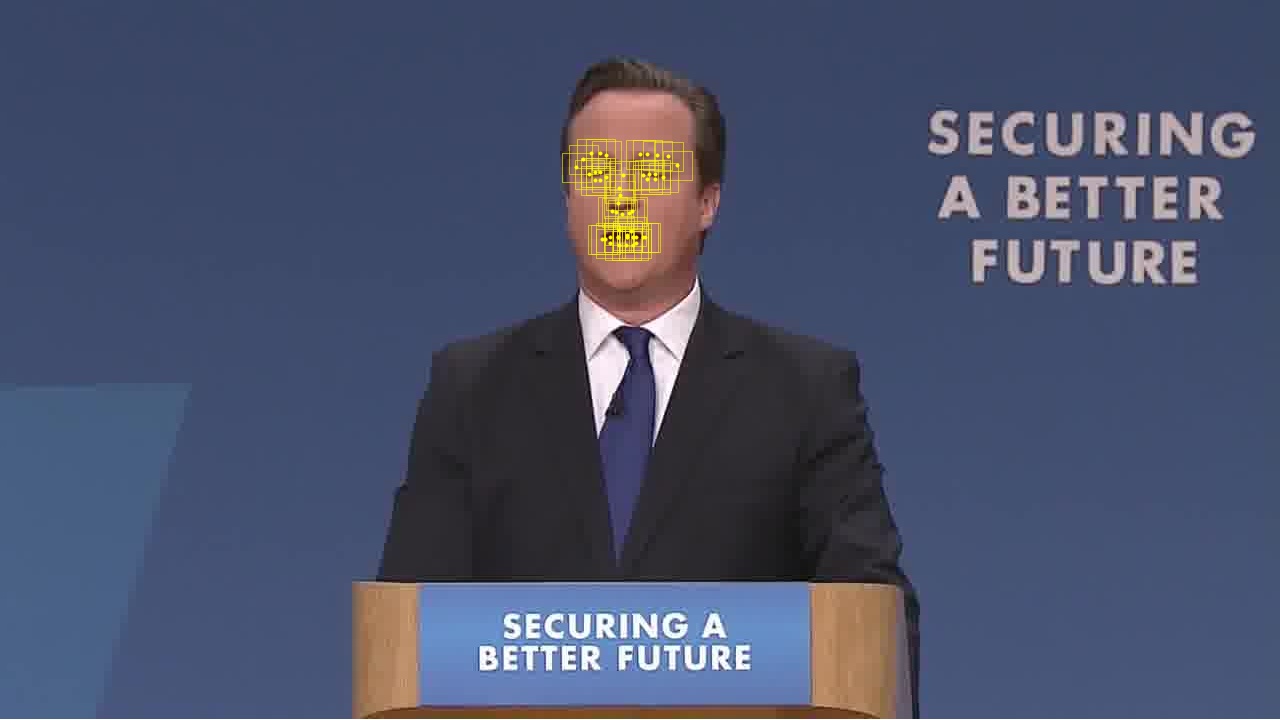}
    \includegraphics[width=0.22\linewidth, trim={15cm 13cm 15cm 0cm}, clip]{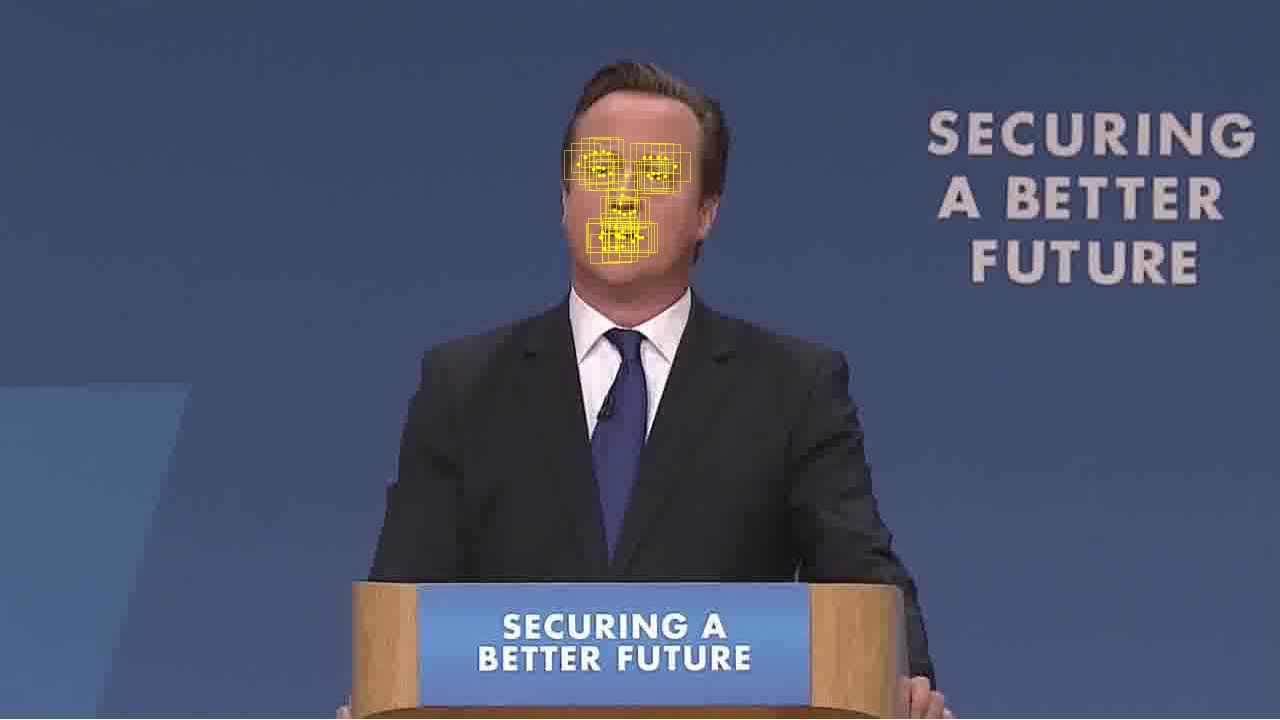}
    \includegraphics[width=0.22\linewidth, trim={15cm 13cm 15cm 0cm}, clip]{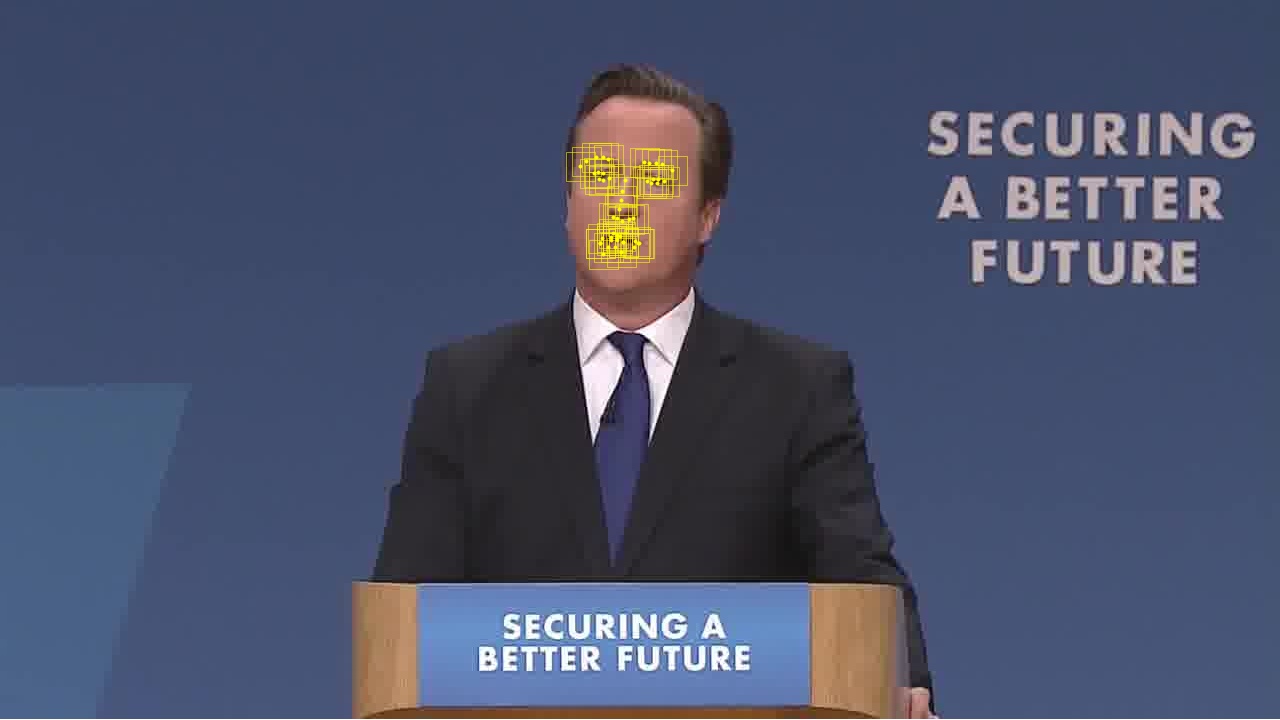}
    \includegraphics[width=0.22\linewidth, trim={15cm 13cm 15cm 0cm}, clip]{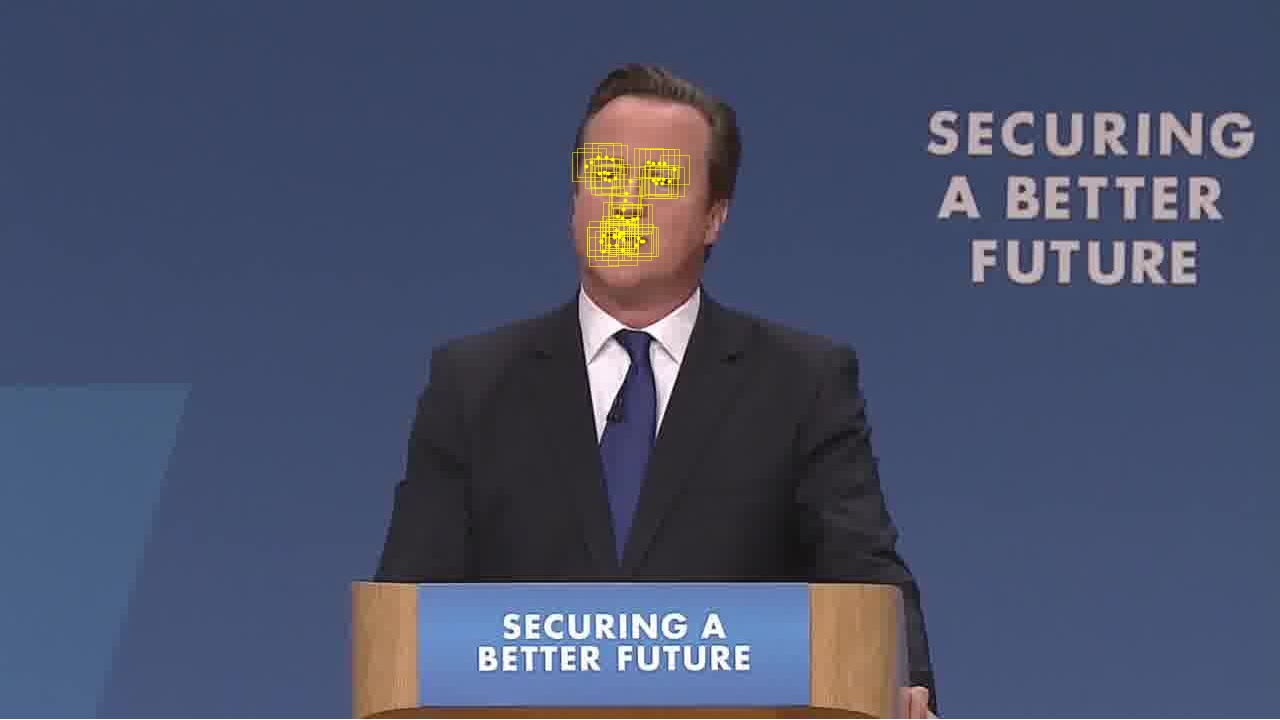}
    \includegraphics[width=0.22\linewidth, trim={15cm 13cm 15cm 0cm}, clip]{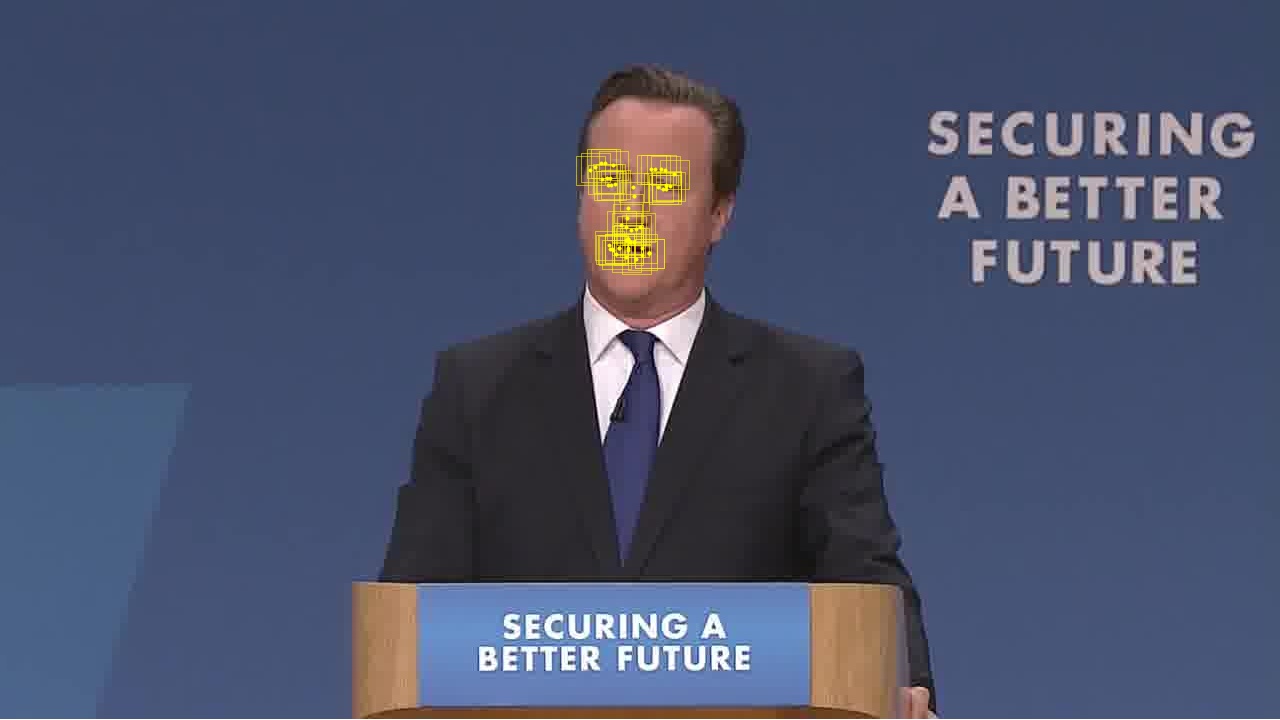}
    \includegraphics[width=0.22\linewidth, trim={15cm 13cm 15cm 0cm}, clip]{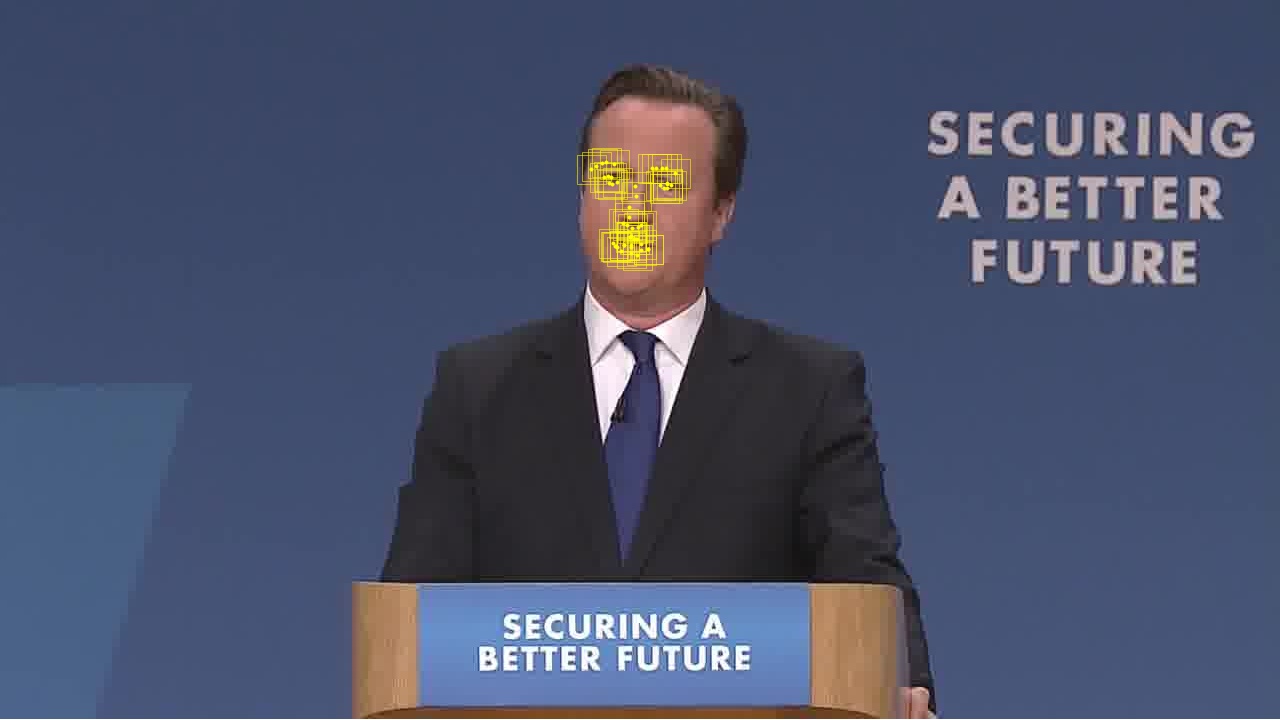}
    \includegraphics[width=0.22\linewidth, trim={15cm 13cm 15cm 0cm}, clip]{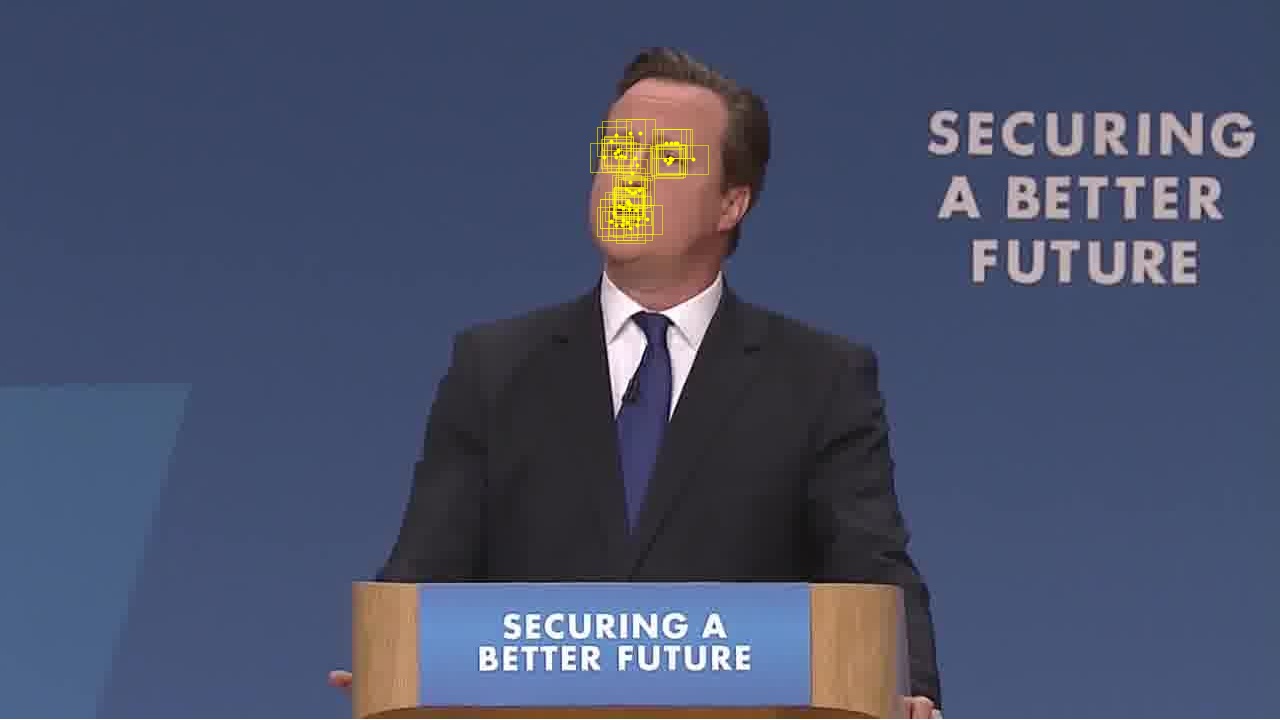}
    \includegraphics[width=0.22\linewidth, trim={15cm 13cm 15cm 0cm}, clip]{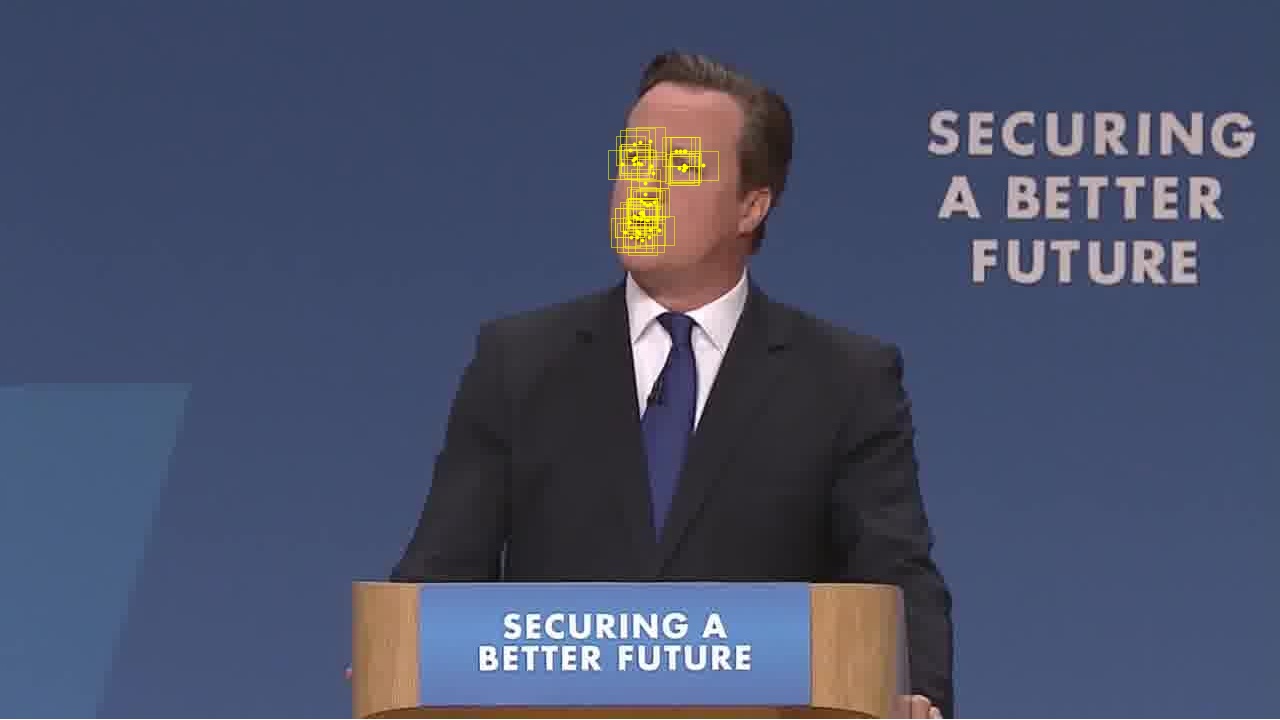}
    \includegraphics[width=0.22\linewidth, trim={15cm 13cm 15cm 0cm}, clip]{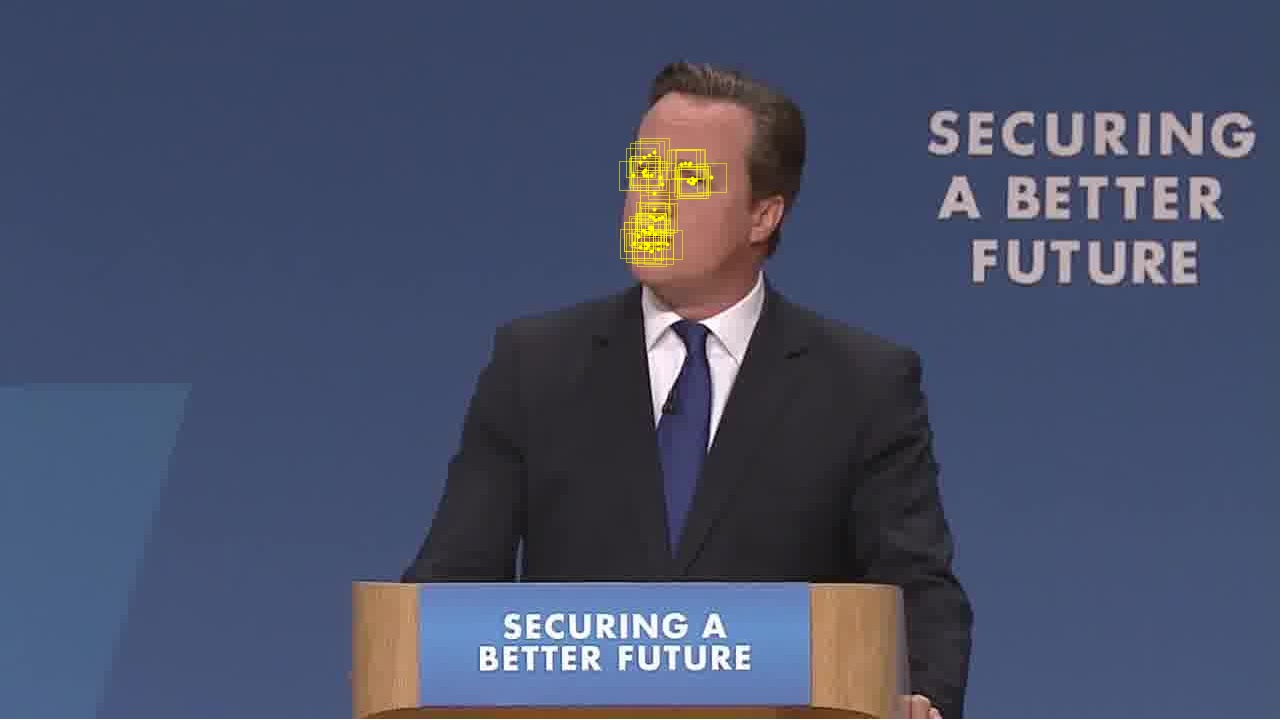}
    \includegraphics[width=0.22\linewidth, trim={15cm 13cm 15cm 0cm}, clip]{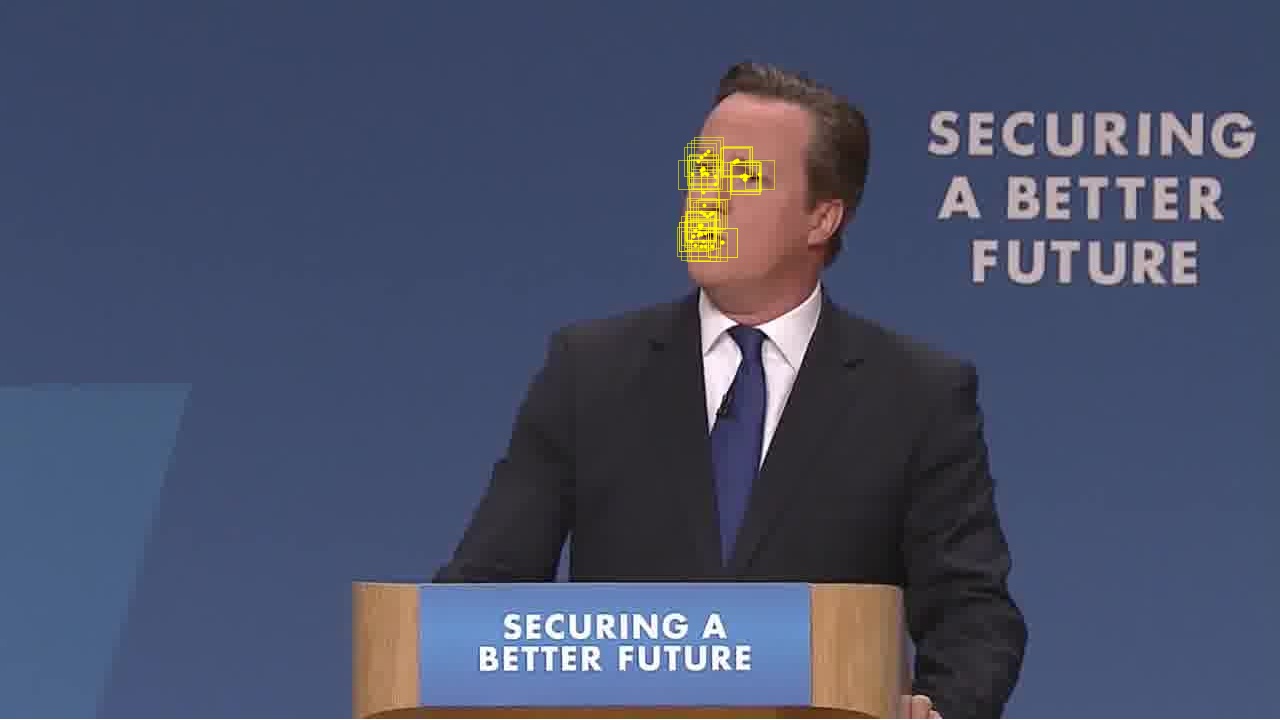}
    \includegraphics[width=0.22\linewidth, trim={15cm 13cm 15cm 0cm}, clip]{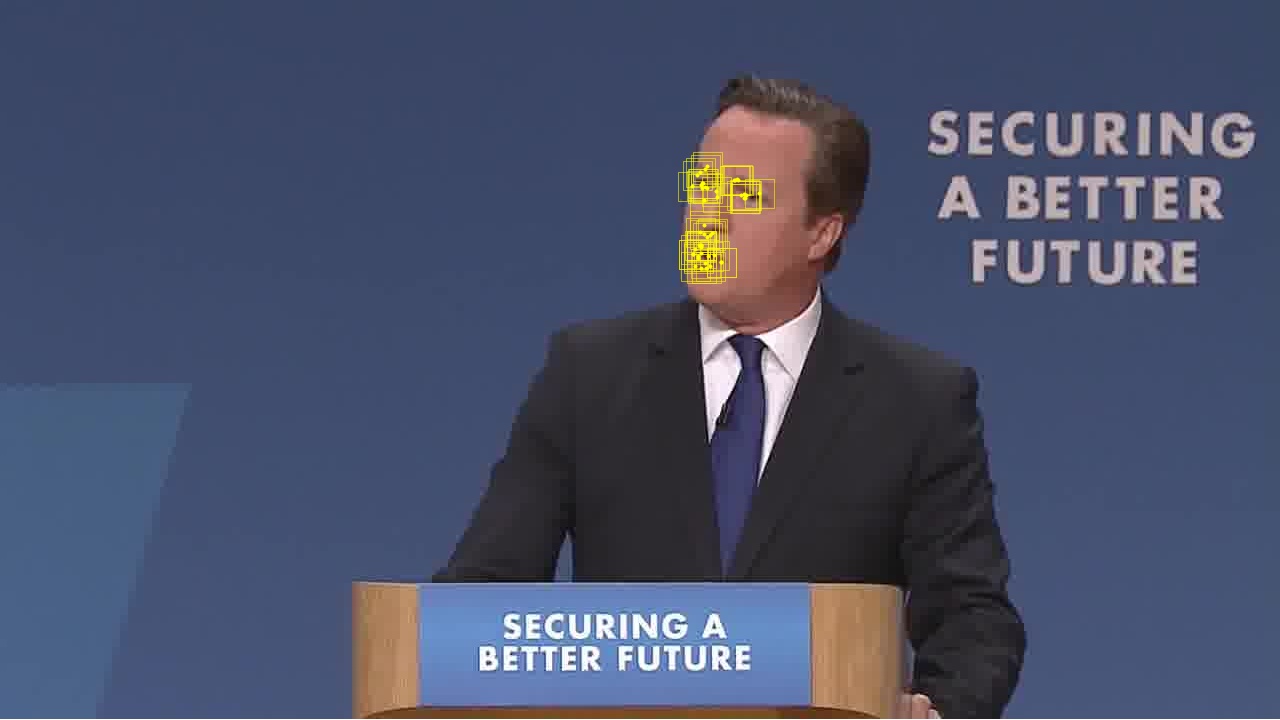}
    \caption{Qualitative results on a sample sequence from the 300-VW dataset.}
    \label{fig:results_300vw}
\end{figure}

\section{Experiments}
\subsection{Implementation Details}
We use first four convolutional layers from VGG-16 \cite{Simonyan2014} with pre-trained weights as our feature extraction part to obtain the image features. To maintain the same resolution with the input images without losing the receptive fields, we remove the pooling layers and increase the dilation of the convolutional layers accordingly. We randomly sample 50 landmarks per pair from input images as the initial landmarks during training. We train the proposed cycle LK network for 20 epochs with a batch size of 1. The Adam optimization is used with an initial learning rate of 0.0001, betas of 0.9 and 0.999, weight decay of 0.000001. Basic data augmentation including scaling, horizontal flipping, a rotation is applied to the input images. All models are implemented in PyTorch \cite{Pytorch}.

\subsection{Evaluation Metrics}     \label{sec:eval}
One pass evaluation (OPE) is used to evaluate the performance of tracking. We run the tracker throughout a test sequence with initialization
from the ground truth position in the first frame. We then measure the average pixel distance error in every frame as $E_d$.
Whenever the tracking for one landmark fails in one image, i.e., $E_d > t$, where $t$ is a threshold, we re-initialize this landmark from the ground truth at the failed frame. The overall performance of the tracker is the mean $E_d$ over all frames and successful rate (i.e., how frequently the tracker does not fail).

The aforementioned error term $E_d$ can be defined differently according to the existence of ground truth of landmarks. While evaluating on THUMOS dataset where we do not have ground truth of landmarks, we define the re-projection error $E_r$ to represent the error,
\begin{equation}
E_r = ||\textbf{x}_T-\textbf{x}_T'||^2
\end{equation}
Where $\textbf{x}_T$ and $\textbf{x}_T'$ is the initialization and backward estimation respectively. When evaluating on dataset where we do have the annotations, we can define the forward error metrics $E_f$,
\begin{equation}
E_f = ||\textbf{x}_G-\textbf{x}_I||^2
\end{equation}
Where $\textbf{x}_G$ is the ground truth locations and $\textbf{x}_I$ is the forward locations estimated by the tracker. In addition, the success rate $S_r$ is defined as 
\begin{equation}
    S_r = \frac{N_s}{N}
\end{equation}
Where $N_s$ is the total number of landmarks successfully tracked on all frames, $N$ is the total number of landmarks on all frames. In the following experiments, we use $t = 0.5$ pixel as the threshold.

\subsection{Qualitative Results}
Qualitative results of sample sequences on THUMOS and 300-VW dataset are shown in Figure \ref{fig:results_thumos} and \ref{fig:results_300vw}. The bounding boxes around the landmark denote the template patch used in Inverse Compositional Lucas-Kanade layer.

\subsection{Quantitative Results}
\vspace{-0.2cm}

\begin{table*}[ht]
\centering
\begin{supertabular}{|c|c|c|c|c|c|c|} \hline
	\multirow{2}{*}{Method}             & \multicolumn{2}{c|}{Basketball-g06-c02}     & \multicolumn{2}{c|}{ApplyEyeMakeup-g04-c02} & \multicolumn{2}{c|}{BaseballPitch-g12-c04}  \\ 
                                        & $E_r$  & $S_r$ &       $E_r$ & $S_r$ &       $E_r$ & $S_r$ \\\hline
      Lucas-Kanade \cite{baker2004lucas}    & 0.231   & 97.1\%       & 0.531     & 95.2\%     & 0.443     & 96.2\%    \\
      CyLKs                             & 0.031   & 99.5\%       & 0.101     &   98.0\%    & 0.151     &   97.7\%   \\
      \hline
\end{supertabular}
\caption{Comparison of Lucas-Kanade tracking and our CyLKs on sequence Basketball, ApplyEyeMakeup, and BaseballPitch from THUMOS dataset.}
\label{table:THUMOS}
\end{table*}

\begin{table*}[ht]
\centering
\begin{supertabular}{|c|c|c|c|c|c|c|c|c|c|} \hline
	\multirow{2}{*}{Method}             & \multicolumn{3}{c|}{seq-3}     & \multicolumn{3}{c|}{seq-18} & \multicolumn{3}{c|}{seq-19}  \\ 
                                        & $E_r$ & $E_f$ & $S_r$ &       $E_r$ & $E_f$ & $S_r$  &       $E_r$ & $E_f$ & $S_r$ \\\hline
      Lucas-Kanade \cite{baker2004lucas}    & 0.432   & 4.816  & 91.2\%     & 0.567     & 4.446     & 92.1\% &      0.832 & 4.176 & 91.9\%\\
      CyLKs                             & 0.187   & 3.525 & 93.9\%          & 0.102   & 4.232     & 92.3\% &       0.177 & 4.886 & 91.1\% \\
      \hline
\end{supertabular}
\caption{Comparison of Lucas-Kanade tracking and our CyLKs on sequence $3$, $18$, and $19$ from 300-VW dataset.}
\label{table:300VW}
\end{table*}

The quantitative results on THUMOS and 300-VW dataset are shown in Table \ref{table:THUMOS} and \ref{table:300VW}. Three error metrics $E_r$, $E_f$, and $S_r$ are used for evaluation. In THUMOS, we only evaluate the error metrics of $E_r$ and $S_r$ as we do not have landmark annotations on for the dataset. We evaluate $3$ sample sequences on these datasets respectively. From the results, we show that the re-projection error $E_r$ (i.e., the difference between backward tracking with the initialization) is substantially lower compared to the Lucas-Kanade algorithm. This demonstrates the cycle loss can help the network learn the transformed features on which we can perform bi-directional tracking. In other words, we can obtain the same results by either tracking from $t-1$ to $t$ frame or from $t$ to $t-1$ frame. Also, we observe that the forward error of CyLKs is lower than Lucas-Kanade by a large margin in most sequences. This demonstrates that the patch loss does enforce the tracked patch in forward pass to be close to the template patch. In brief, we show that, without using any annotations from manual labeling, we can achieve a significant improvement in both forward tracking and re-projection.

\vspace{-0.2cm}
\section{Future Work}
\vspace{-0.2cm}

In this work, we propose a Cycle Lucas-Kanade (CyLKs) Network to achieve landmark tracking in an unsupervised way.
Using the proposed cycle loss and patch loss, we can achieve a significant improvement on landmark tracking without using any manual labels.
With the success of CyLKs, it is interesting to show how the learned model can improve existing LK-based tracking systems. In other words, we can use off-the-shelf CyLKs to extract features for LK-based tracking systems instead of using raw images. Another interesting direction is to extend the work from landmark tracking to general object tracking by replacing current 2 DoF (degree of freedom) translation parameters with 6 DoF affine transformation.


{\small
\bibliographystyle{ieee}
\bibliography{egbib}
}

\end{document}